\documentclass[10pt,twocolumn,letterpaper]{article}
\usepackage{iccv}
\usepackage{times}
\usepackage{epsfig}
\usepackage{graphicx}
\usepackage{amsmath}
\usepackage{amssymb}
\usepackage{capt-of,etoolbox}
\usepackage{comment}
\usepackage{comment}
\usepackage{multirow}
\usepackage{xcolor}
\usepackage{lipsum}
\usepackage{graphicx}
\usepackage{caption} 
\usepackage{subcaption}
\usepackage{authblk}
\usepackage{bigfoot}
\usepackage[bottom]{footmisc}

\usepackage[pagebackref=true,breaklinks=true,colorlinks,bookmarks=false]{hyperref}

\iccvfinalcopy 

\begin{document}

\title{ISETHDR: A Physics-based Synthetic Radiance Dataset for High Dynamic Range Driving Scenes}

\author[1]{Zhenyi Liu}
\author[2]{Devesh Shah*} 
\author[1]{Brian Wandell}

\affil[1]{Stanford University, Stanford, CA, USA \quad \textsuperscript{2}Ford Motor Company, Dearborn, MI, USA}

\affil[ ]{\tt\small \{zhenyiliu, wandell\}@stanford.edu}



\maketitle



\begin{abstract}
This paper describes a physics-based end-to-end software simulation for image systems. We use the software to explore sensors designed to enhance performance in high dynamic range (HDR) environments, such as driving through daytime tunnels and under nighttime conditions. We synthesize physically realistic HDR spectral radiance images and use them as the input to digital twins that model the optics and sensors of different systems. This paper makes three main contributions: (a) We create a labeled (instance segmentation and depth), synthetic radiance dataset of HDR driving scenes. (b) We describe the development and validation of the end-to-end simulation framework. (c) We present a comparative analysis of two single-shot sensors designed for HDR. We open-source both the dataset and the software: \href{https://github.com/ISET/isethdrsensor}{ISETHDRSensor}.
\end{abstract}
\def\thefootnote{*}\footnotetext{Work performed while at Ford Motor Company.}

\begin{figure*}[t]
    \centering
    \includegraphics[width=0.99\textwidth]{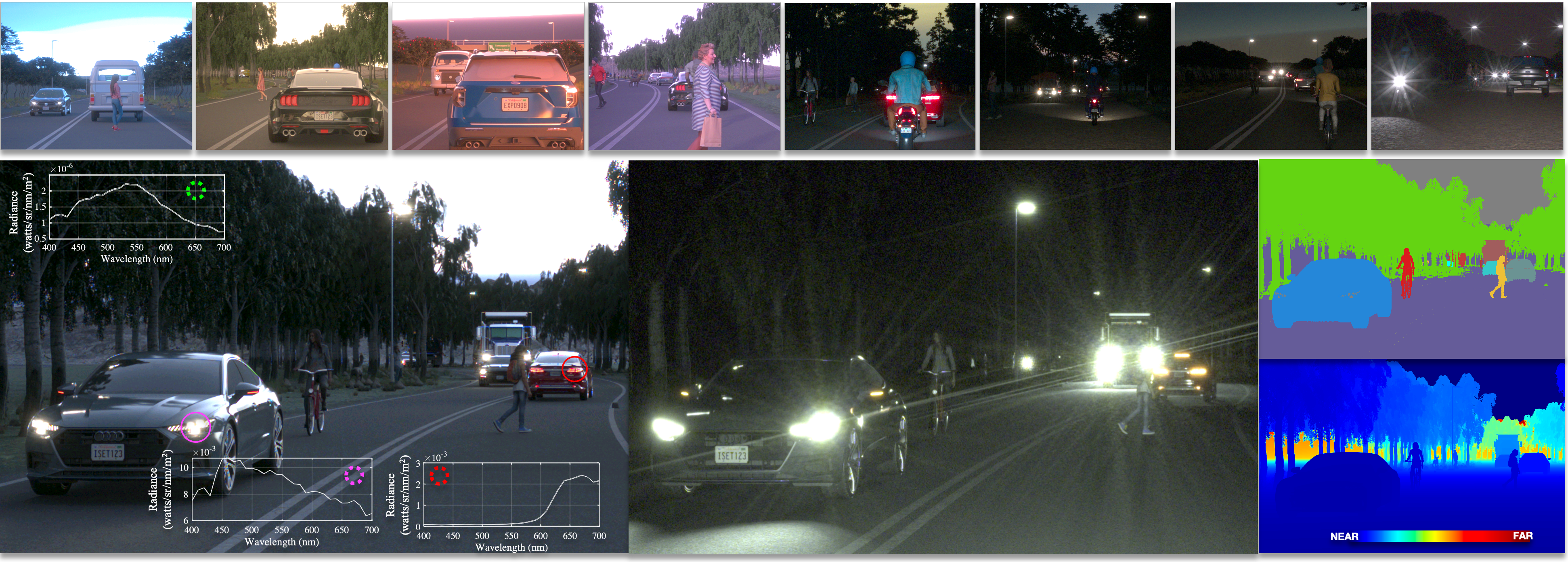}
    \caption{\textbf{Overview.} A physics-based end-to-end image systems simulator was developed to synthesize spectral radiance of complex scenes and to model both image formation and image sensor designs. This work introduces a driving scene dataset (top row) comprising renderings illuminated by distinct light groups (sky, headlights, streetlights, and other sources). Daytime and nighttime scene spectral radiance were simulated by linearly combining these light group renderings (bottom left and middle images, respectively). Pixel-level object labels and depth maps are provided for each scene (right).}
    \label{fig:openfig}
\end{figure*}

\section{Introduction}
Acquiring, analyzing, and displaying high dynamic range images, such as the nighttime driving images, poses significant challenges (see Figure \ref{fig:openfig}). First, these images often contain very bright, localized, and directional light sources (headlamps, streetlights, traffic lights). The  radiance level of these sources can be five orders of magnitude higher than the radiance level in nearby dark regions; the dynamic range of these scenes exceeds the range that can be faithfully captured by most single-shot image sensors. 

Second, the absolute intensity level in large regions of nighttime driving scenes (say, 1-10 $cd/m^2$) is such that sensors with pixels 1-3 microns in size, optics with an f-number around 4, and a 60 Hz (16 ms) exposure duration only capture a few dozen photons per pixel. Even for an ideal sensor, the Poisson noise at this level severely limits the image contrast-to-noise ratio. 

Third, the light level from optical flare near the bright sources can be large, compared to the light level in nearby parts of the image. Flare - whose properties are determined by the aperture shape, scratches and dust, and inter-reflections between optical elements - spreads light and creates image contrast that is not present in the scene. Such flare photons are present in daytime images, too, but they are less visible when they are superimposed upon a higher background (Figure~\ref{fig:openfig}). Lens flare can introduce image artifacts, reduce contrast, and obscure important details such as a pedestrian in front of a bright light source. These limitations create dangerous conditions for driving.

This paper describes image system designs intended to confront the challenges of acquiring high quality HDR driving scenes with imaging sensors. This paper contributes a dataset of 2000 semantically labeled spectral scenes, organized into \textit{light groups} - collections of four spectral radiance maps that can be combined in many ways to simulate different lighting conditions (Figure~\ref{fig:openfig}). This paper also describes open source, free, software tools for a physics-based, quantitative end-to-end simulation (digital twin) from scene spectral radiance to processed image sensor data of static HDR driving scenes. The spectral radiance data can be input to the image systems simulation software. The simulation processes the scene radiance through a lens model, including varying degrees and type of flare, to calculate the sensor irradiance. It then processes the irradiance through a sensor model. Finally, we simulate and compare the performance of two single-shot, high dynamic range, image sensor designs.

\section{Related Work}
We describe related work that (a) creates and analyzes high dynamic range driving images and (b) simulates flare. 

\subsection{HDR Nighttime driving data}

The challenge of measuring and labeling real HDR nighttime driving scenes has driven several research groups to develop methods for creating synthetic nighttime driving images. One recent approach estimates the pixel-wise illumination (RGB) in a daytime image and then relights it with pixel-wise nighttime illumination \cite{Punnappurath2022-day2night}. Another approach employs domain adaptation techniques to transform labeled daytime images into nighttime images \cite{Dai2018-semanticsegDayNight, Bui2020-pb, Sakaridis2022-DANighttimeSeg, Xu2021-nightSegCurriculum, Wu2023-DANNet}. For example, one method progressively trains a transition from daytime to twilight to nighttime using a sequence of images. Another method uses labeled daytime images that are roughly aligned with unlabeled nighttime images. A third example leverages adversarial networks, such as CycleGAN, to learn the conversion of collections of daytime images into unpaired nighttime images \cite{Lin2021-d2n-IEEEIntellTransport, Zhu2017-cycleGAN}.

While these methods generate RGB images that appear visually plausible, they do not account for image system components, limiting their utility for image system design.

There are multiple databases that contain RGB images of nighttime driving scenes \cite{Wang2023-nighttimedata,Sakaridis2019-darkzurich, Sun2019-NightSemSeg-ZJUdata,Yu2018-Berkeley-BDD}. Most of these data were acquired by conventional cameras; others were synthesized by commercial software. Some of the nighttime driving datasets also include semantic labels.

The datasets do not include quantified scene spectral radiance: The labeled scene spectral radiance data, which we provide in light groups, is the only quantified, labeled, dataset we know of. The quantitative representation of the scene spectral radiance is essential for the image system design evaluations we carry out in this paper. 

\subsection{Optical flare simulation}
Wu et al. \cite{Wu2020-flare-bl} describe a model that accounts for two of the three key components of flare: aperture shape and surface imperfections (dust and scratches). Working in the RGB domain, they simulate flare from light sources and then add the RGB flare images into standard RGB images at locations with high-intensity light sources. A similar approach is employed by Dai et al. \cite{Dai2022-flare7k, Dai2023-flare7k++}, who provide a database of 7,000 RGB flare images generated using commercial software.

While this method creates images with the visual appearance of flare, it does not include a physical model of the scene or the image system. As a result, the added flare may not accurately match the intensity of the actual light source. Achieving such accuracy is nearly impossible in the RGB domain, especially in nighttime driving scenes, where the intensity of light sources can be four to six orders of magnitude greater than that in dark regions. Deriving the true value from conventional RGB images is either difficult or impossible.

Another limitation of this approach is that flare is not inherently additive. For shift-invariant optics, flare is modeled by the point spread function, which depends on the aperture size and shape, surface imperfections, and other wavefront aberrations of the optics (see Methods). Adding an RGB image of flare to an existing RGB image fails to capture the convolutional nature of the flare calculation or the true spatial extent of the light source. This additive approach is particularly inaccurate when the bright light is an area source, such as a headlight or street lamp.

\subsection{Applications}
Several applications can benefit from flare and nighttime driving datasets. One application is to train flare removal and flare addition networks for consumer photography \cite{Wu2020-flare-bl}. A second application is to solve difficulties in obtaining semantic labels of nighttime images for autonomous vehicle planning and control system \cite{Punnappurath2022-day2night, Dai2018-semanticsegDayNight, Sakaridis2022-DANighttimeSeg,Wu2019-night-rf,Wu2023-DANNet}. The labeled RGB images are useful for training networks to provide better information for downstream applications, such as object segmentation in nighttime driving images \cite{Qiao2021-flare, Zhou2023-flareremoval-ICCV, Dai2022-flare7k}. 

The application in this paper is to evaluate single-shot, HDR sensors designed to provide high quality sensor information under HDR lighting conditions.

\section{Methods}
\label{sec:Methods}
The fundamental image system simulation methods are detailed in previous publications \cite{Farrell2012-digcamsimApOp, Liu2019-softprototype-ICCV}. These simulations leverage physically-based ray tracing \cite{pharr-book}, utilizing high-quality assets, spectral light sources, and materials. The accuracy of the simulations has been validated in several studies \cite{lyu2021validation, Goossens-RTF, Farrell2008-sensorcalandsim, Chen2009-isetvalidation}. The open-source software and validation data are available in a collection of repositories under the GitHub ISET organization, including the core repositories \cite{iset3dsoftware, isetcamsoftware}. In this paper, we describe the new features we developed to simulate nighttime driving scenes and provide scripts to generate several of the figures included in the shared code: \href{https://github.com/ISET/isethdrsensor}{ISETHDRSensor}.

\subsection{HDR scene simulation}
We used Roadrunner \cite{roadrunner-mainpage} to create realistic road simulations. This software integrates road descriptions from various open-source standards (e.g., OpenStreetMap, OpenScenario, OpenDrive). We developed 25 different types of base roads to ensure a diverse range of road environments.

To model vehicles, vulnerable road users, and the surrounding environment, we created a comprehensive collection of 3D assets, including over 80 vehicles (e.g., cars, buses, trucks), 30 pedestrians, 35 cyclists with bicycles or motorcycles, and more than 70 trees, along with grass, rocks, and various animals. Additionally, we calibrated a set of spectral power distributions for common street lighting and headlamps to achieve realistic illumination. We also collected high dynamic range environmental lights (sky maps) that represent typical spectral power distributions at different times of day, from morning to night.

We assembled a diverse set of scenes using these assets, as detailed in our previous work \cite{Liu2020-generalization, Liu2021-depth-radiance, lyu2021validation}. Below, we describe the new features specifically added to address the quantitative simulation of nighttime scenes.

\subsubsection{Light source simulation}

\begin{figure}
    \centering
    \includegraphics[width=0.45\textwidth]{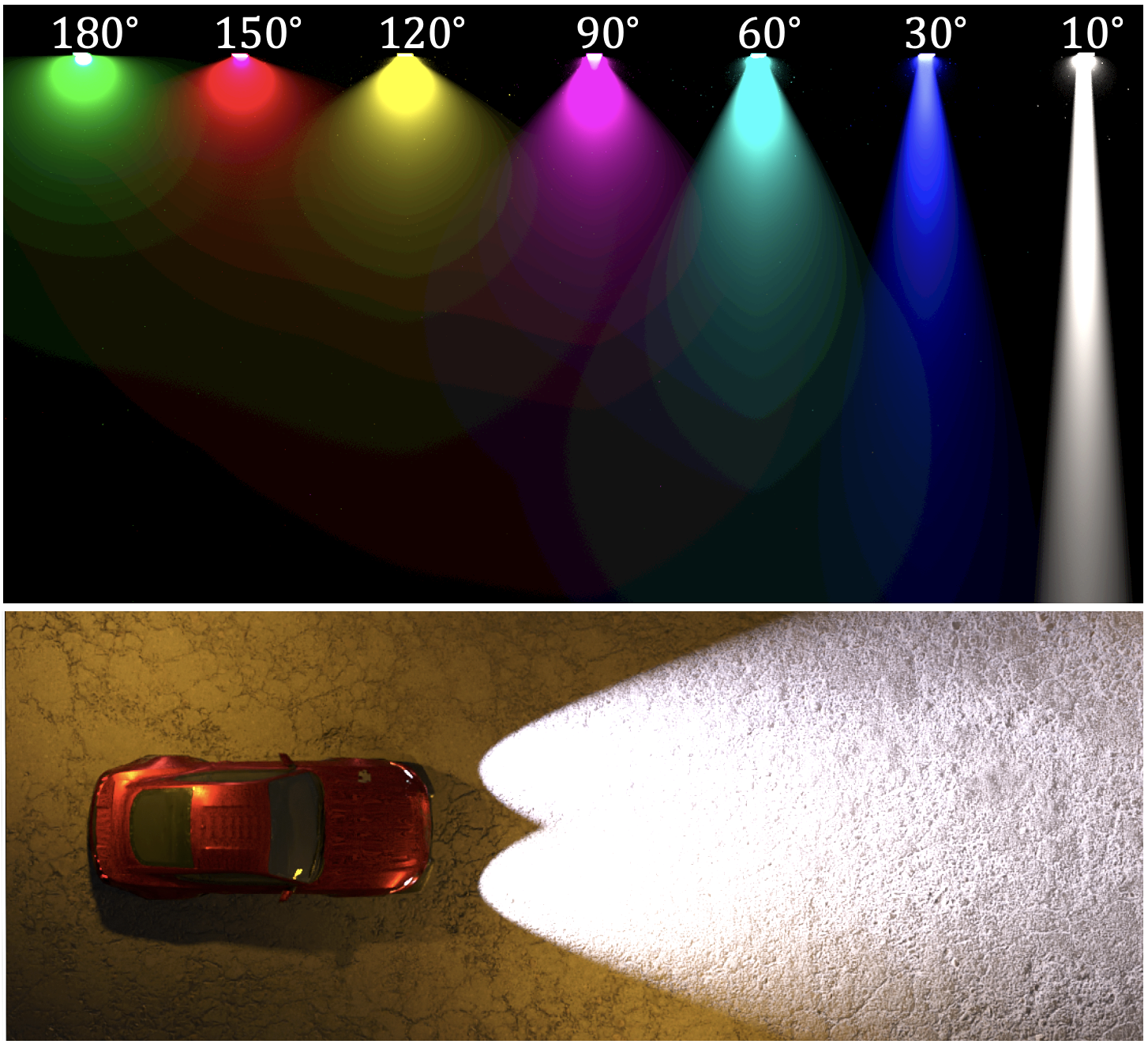}
    \caption{\textbf{Beam angle control.} To accurately simulate various light sources such as headlights, taillights, and streetlights, we extended the PBRT arealight model to include controllable beam angle parameters (top). The bottom image demonstrates the limited beam angle area lights for car headlights.}
    \label{fig:beamAngle}
\end{figure}

\begin{figure*}[ht]
    \centering
    \includegraphics[width=0.99\textwidth]{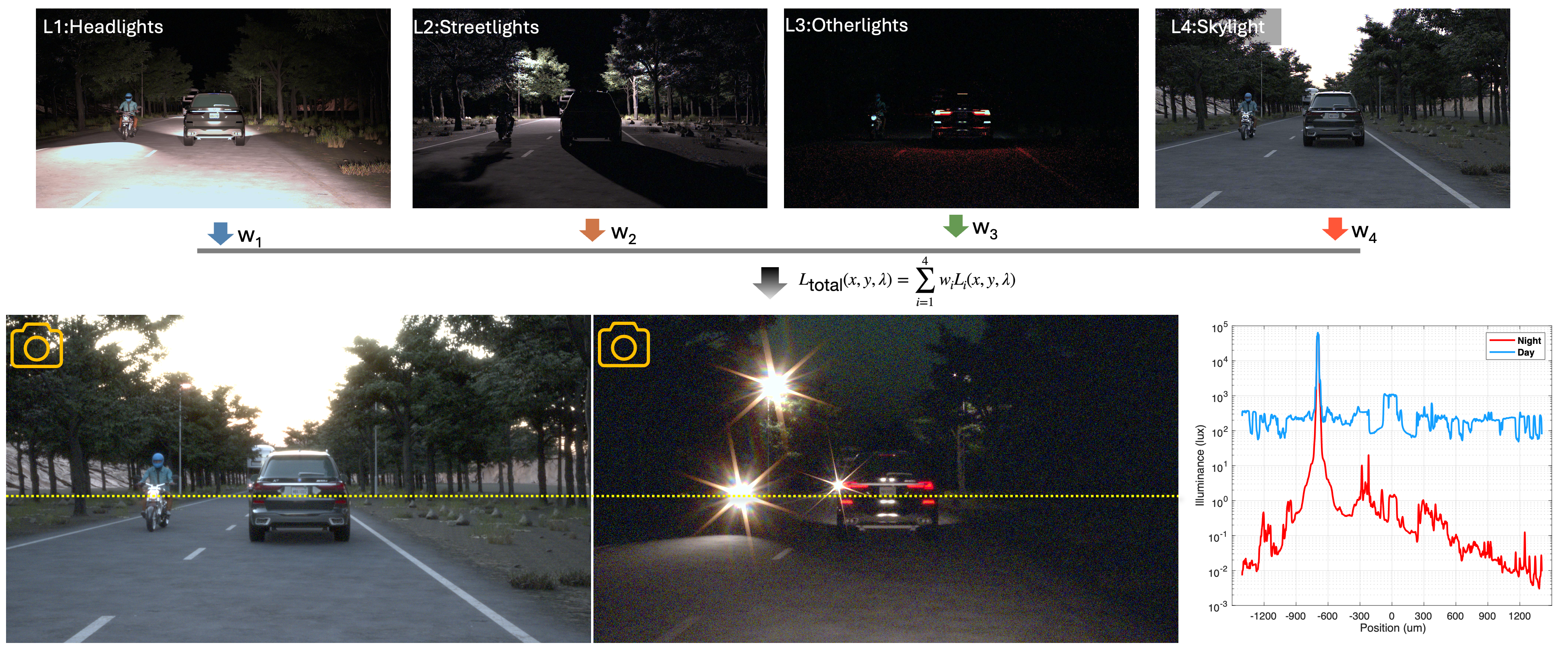}
    \caption{\textbf{Scene light groups.} The dataset comprises 2000 scenes, each defined by four spectral radiance maps representing illumination by the sky, headlights, streetlights, and other light sources (e.g., tail lights, bicycle lights). To simulate various lighting conditions, the four maps are combined with different weights. For example, a daytime scene (left) has a bright sky and headlights, while a nighttime scene (right) has a darker sky with prominent headlights and streetlights. Using a lens model incorporating aperture and scratch effects (but excluding inter-reflections), scene radiance is converted to sensor irradiance. The graph on the right illustrates the illumination profile across a horizontal line. Notably, headlight intensity remains constant between day and night, while reduced skylight lowers image contrast in darker areas. The software includes tools to select the appropriate weights for achieving the desired dynamic range and low-light conditions.(lightGroupDynamicRangeSet.m).
    }
    \label{fig:lightgroups}
\end{figure*}

We simulate light sources with area lights for several important reasons. Spotlights are sometimes used for simulating headlights and other artificial lights with a narrow spread angle. However, spotlights are emitted from a point, which means the light source surface does not appear in the rendering. Hence, spotlights are not appropriate for simulating nighttime driving scenes with visible headlights or street lights. 

The ray tracing software, PBRT, includes area lights, which are appropriate for this application. The area lights have a surface and the outgoing rays are emitted in the hemisphere pointed to by the surface normal. However, PBRT does not have parameters that control the area light spread, which is necessary for simulating headlights and street lights. To overcome this limitation, we introduced beam angle controls into the PBRT area light model (Figure~\ref{fig:beamAngle}). We incorporated these area light sources into the 3D car models, including headlights, taillights, indicators, and brake lights, to ensure an accurate representation of the car’s lighting conditions in our system. These area lights are also used for street lights.

\subsubsection{Light groups} 
Complex driving scenes include multiple light sources. These lights -- such as sunlight, headlights, and streetlights -- are incoherent sources whose wavefront are in random phase. To simulate these scenes, we can sum the scene radiant energy, rather than combining the wavefront amplitudes and phases as would be required for coherent lights. Thus, the total radiant energy \( L_{\text{total}} \) of a scene at a point \((x, y)\) and for a specific wavelength \(\lambda\) can be computed as the weighted sum of the contributions from the different sources \( L_i \).

\begin{equation}
    L_{\text{total}}(x, y, \lambda) = \sum_{i=1}^{N} w_i L_i(x, y, \lambda)
\end{equation}

We use this observation to implement an efficient method for simulating diverse lighting conditions of a scene. By decomposing the scene into distinct light groups (sky, headlights, streetlights, and other sources) and rendering spectral radiance maps for each (Figures \ref{fig:openfig} and \ref{fig:lightgroups}), we can reconstruct illumination at various times of day without additional ray tracing. Each driving scene in our public dataset was rendered four times, once for each light group. To simulate different lighting scenarios, these radiance maps are linearly combined with weights adjusted to match target illuminance levels. For instance, sky map intensity varies significantly between day and night (approximately 3,000 cd/m² to several orders of magnitude lower), while headlights, streetlights, and tail lights are active primarily during twilight and nighttime (often exceeding 6,000 cd/m²). Additionally, software tools are included to control the dynamic range and overall illumination of the composite scenes.

\subsection{Image system simulation}
\subsubsection{Optics simulation}

\begin{figure}[t]
    \centering
    \includegraphics[width=0.45\textwidth]{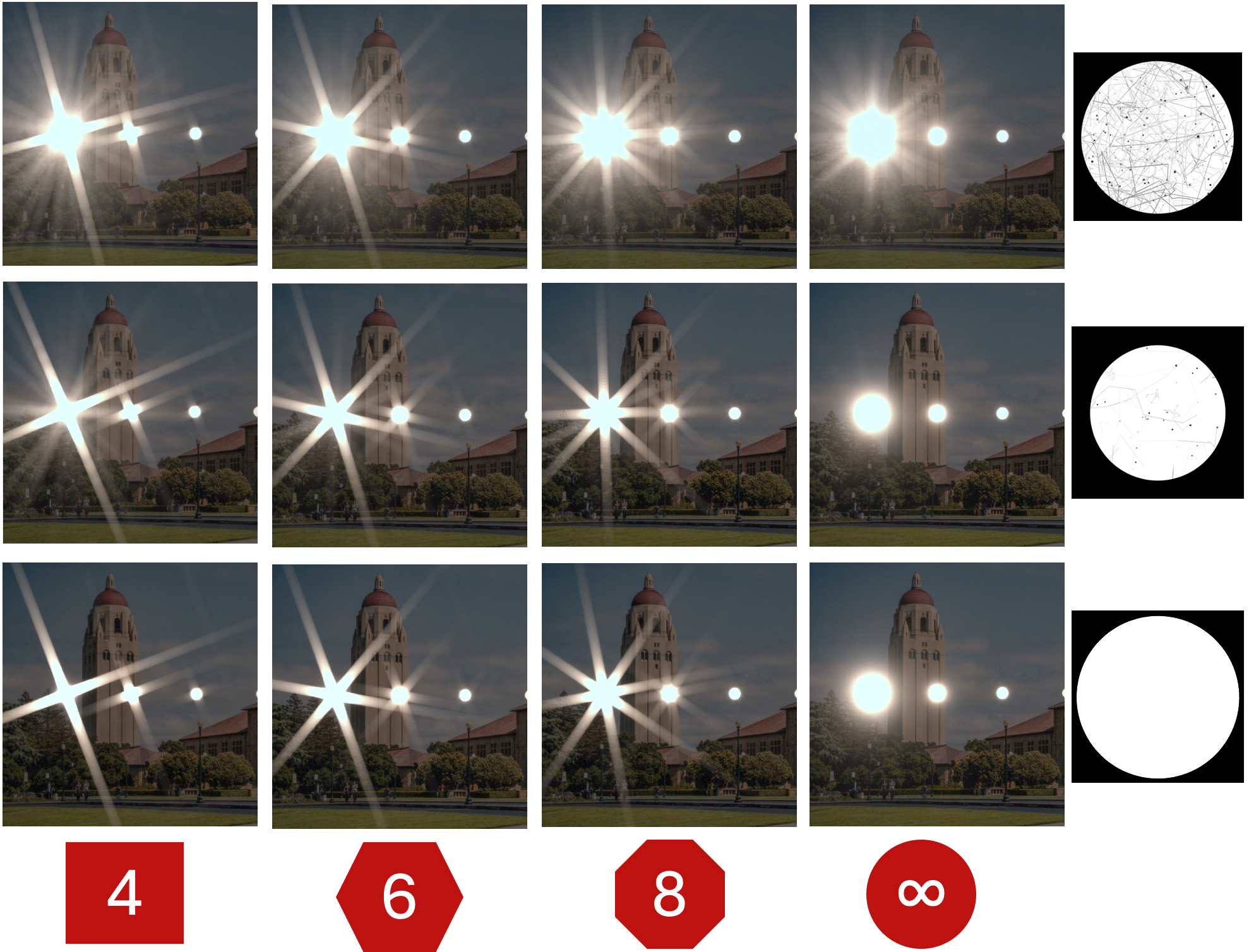}
    \caption{\textbf{Flare model.} The figure depicts a series of simulated scenes featuring an array of bright lights, resembling headlights, with a dark image in the background. The bright light intensities each step down by a factor of 10 across the image. Each scene was rendered using distinct flare parameters. The number of aperture blades increases from four (leftmost column) to a circular aperture (rightmost column). The density of simulated dust and scratches varies from high (top row) to minimal (bottom row).}
    \label{fig:controlledLensFlare}
\end{figure}

We implemented a physics-based simulation of a major source of flare: scattering. This flare arises from the aperture boundary and lens dust and scratches. Scattering flare causes light to deviate from the designed path, and it can introduce image artifacts called streaks (or spikes). A second type of flare - reflective flare - arises from surface inter-reflections within a multi-element lens system. Modern lens coatings typically reflect less than $1\%$ of the light, and to reach the sensor reflective flare must reflect from at least two surfaces. Consequently, the reflective lens flare from any pair of surfaces is low, typically less than $10^{-4}$ of the light intensity. But in some lens designs there are many surface combinations, and in HDR scenes reflective flare can have significant image contrast.

We modeled scattering flare by combining a lens model and a scattering model into a single optical wavefront \cite{Wu2020-flare-bl}. The lens model was characterized by a wavefront aberration function, $\phi(x,y,\lambda)$, describing the optical performance of the ideal, flare-free lens. The scattering model is defined by a wavelength-dependent apodization function $a(x,y,\lambda)$ which accounted for both the aperture shape and imperfections such as dust and scratches. By multiplying these functions, we created a pupil function that comprehensively modeled the lens and the associated scattering flare. Figure~\ref{fig:controlledLensFlare} presents examples of different flare patterns generated by varying aperture shapes and levels of dust and scratches.

\begin{equation}
    w(x,y,\lambda) = a(x,y,\lambda)*\exp^{i \phi(x,y,\lambda)}.
\end{equation}

The magnitude of the Fourier Transform ($\mathbb{F}$) of this complex function is the point spread function ($P$).

\begin{equation}
    P(x,y,\lambda) = \left | \mathbb{F}(w(x,y,\lambda) \right | ^ 2
\end{equation}

A scene can be rendered into an irradiance at the sensor surface (optical image) using different lens and flare parameters. We first apply a wavelength-by-wavelength convolution of the point spread function with the spectral radiance. We then apply the lens geometric distortion and relative illumination to the optical image spectral irradiance.

\begin{figure}
    \includegraphics[width=0.5\textwidth]{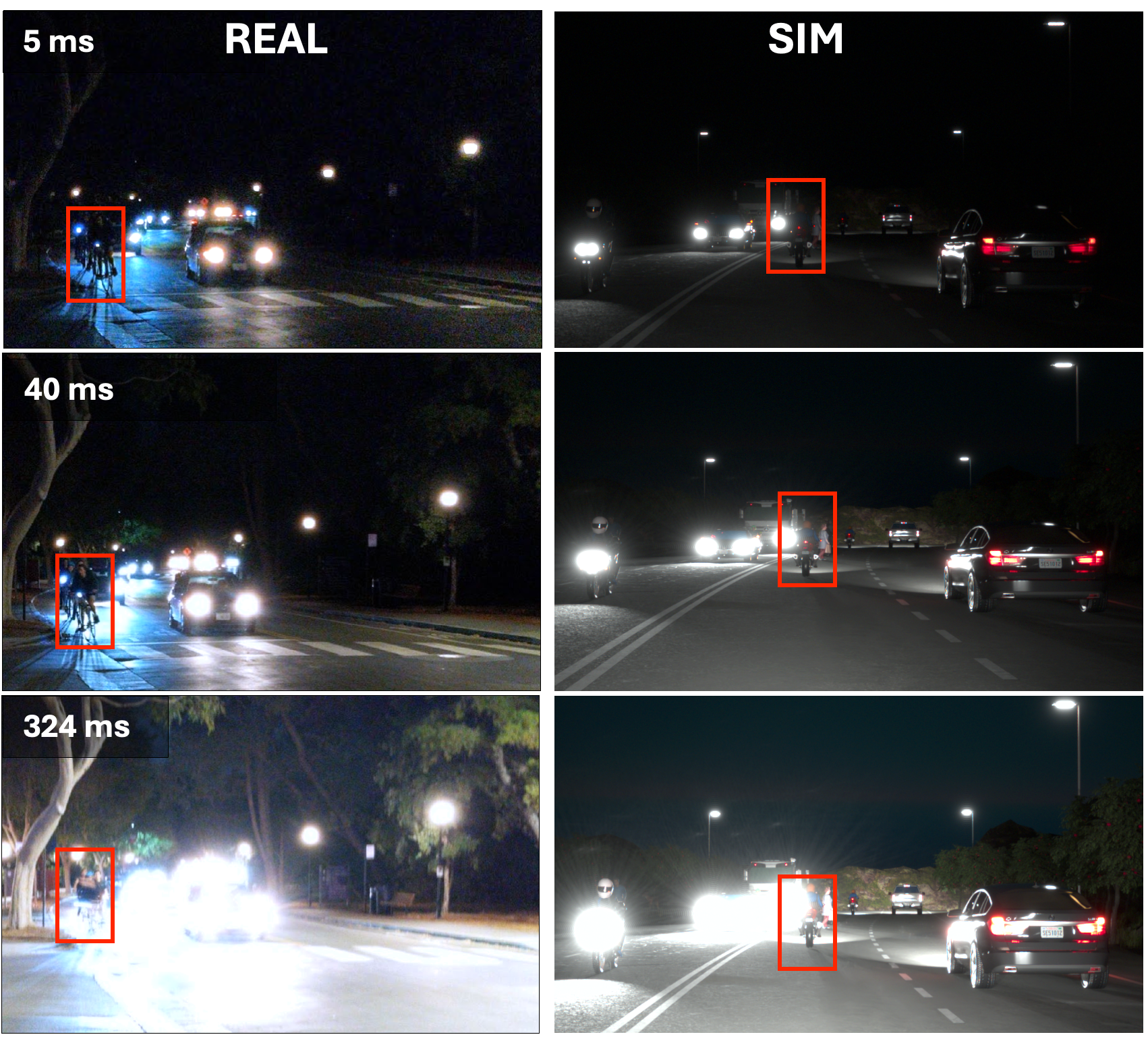}
    \caption{\textbf{Simulated and measured nighttime driving scenes.} \textit{Left:} Images in this column were captured in rapid succession by a Google Pixel 4a, with exposure duration increasing from top to bottom (see inset). \textit{Right:} Images in this column were simulated, using a model of the Google Pixel 4a \cite{Lyu2022-validation}. The spatial extent of the flare, and the corresponding extent of the sensor saturation, are very similar when comparing the two columns. The red boxes outline two vulnerable road users: a cyclist (left) and motorcyclist (right). As exposure duration increases from 5 to 40 ms both cyclists become more visible. At the longest duration, the flare  - arising from headlights behind the cyclists - expands and masks a significant part of these vulnerable road users.}
    \label{fig:flareCompare}
\end{figure}

\subsubsection{Sensor simulation}
We use ISETCam sensor models to convert the optical image irradiance into pixel voltages and digital values.  These models have been validated in controlled laboratory experiments \cite{Lyu2022-validation}. We qualitatively compare the expected nighttime driving images over a range of exposure times in Figure~\ref{fig:flareCompare}. These images, which are rendered using conventional demosaicing and color transformations, illustrate the challenges in using conventional sensors to capture the high dynamic range nighttime driving scenes. They also illustrate the similarity between the measured and simulated images.

\subsection{Dataset}
The ISETHDR  dataset comprises 2000 light groups, each contains four spectral radiance maps, a corresponding depth map, and instance segmentation data. The scenes are country roads flanked by vegetation and with a diverse range of vehicles (cars, buses, trucks) .  The scenes include a variable number and arrangement of vulnerable road users (e.g., people, deer, cyclists) . The scene metadata includes bounding boxes derived from the instance segmentation. 

The light group data serves as a foundation for generating a vast array of labeled images. Spectral radiance maps can be combined to simulate various lighting conditions, while the introduction of different lenses and flare patterns expands the dataset’s diversity.  By applying a wide range of sensor models with varying color filter arrays, pixel sizes, and other parameters, the initial 2000 scenes can be transformed into a substantially larger dataset suitable for training and testing algorithms employed in driving applications.

\section{Experiments}

\begin{figure}[t]
    \centering
    \includegraphics[width=0.95\linewidth]{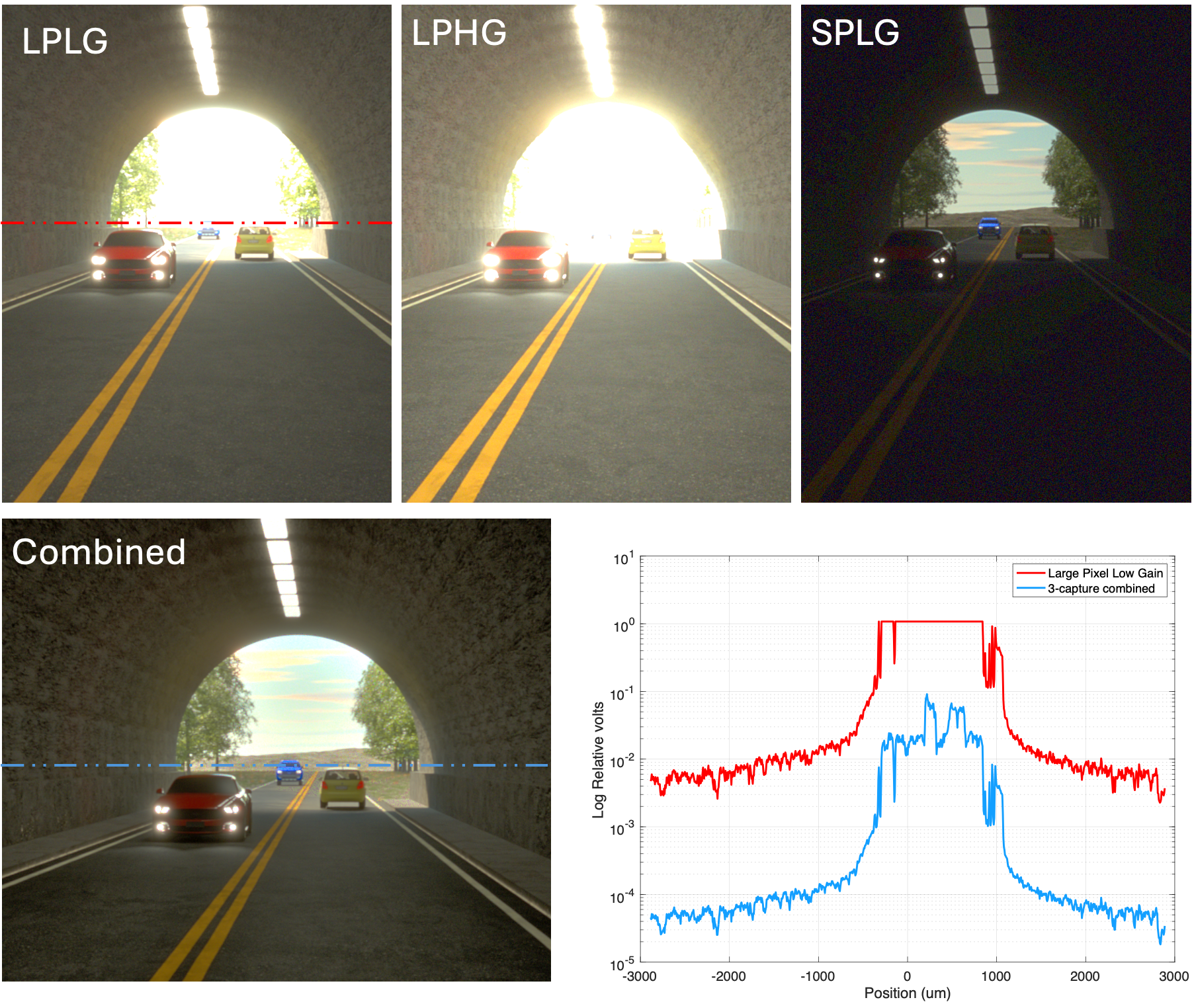}
    \caption{\textbf{A split pixel 3-capture sensor in a tunnel}. We modeled a CMOS image sensor (CIS) with a split pixel, 3-capture, design \cite{Solhusvik2019-split}. Each pixel contains a large and small photodetector. The sensor acquires two images from the large photodetector, one is read with low gain (LPLG) and a second with high gain (LPHG). The third image is acquired using a small photodetector with low gain (SPLG).  Simulated images of these three captures are shown across the top of the figure. The image at the bottom left (Combined) is reconstructed from the three captures. The graph at the lower right shows the log relative voltage from the LPLG sensor (red) and the combined sensor (blue) across a horizontal image line. The saturation of the LPLG data in the tunnel opening is evident; the combined sensor data preserve image contrast across the entire scene. The two curves are displaced vertically from one another for clarity.}
    \label{fig:splitTunnel}
\end{figure}
In the experiments below, we simulate two types of high dynamic range sensors. The first is an automotive sensor with RGB color filters based on a split pixel design described by Omnivision (\cite{Willassen_2015_splitpixel}). The second is an RGBW sensor (also called RGB-clear) described by On Semiconductor. The sensor design and parameters (e.g., pixel size, noise, fill factor, color filter array, etc.) are described more fully below, and the complete set of parameters are provided in the software repository.

\subsection{Split pixel (3-capture) sensor}
\begin{figure}[t]
    \centering
    \includegraphics[width=0.95\linewidth]{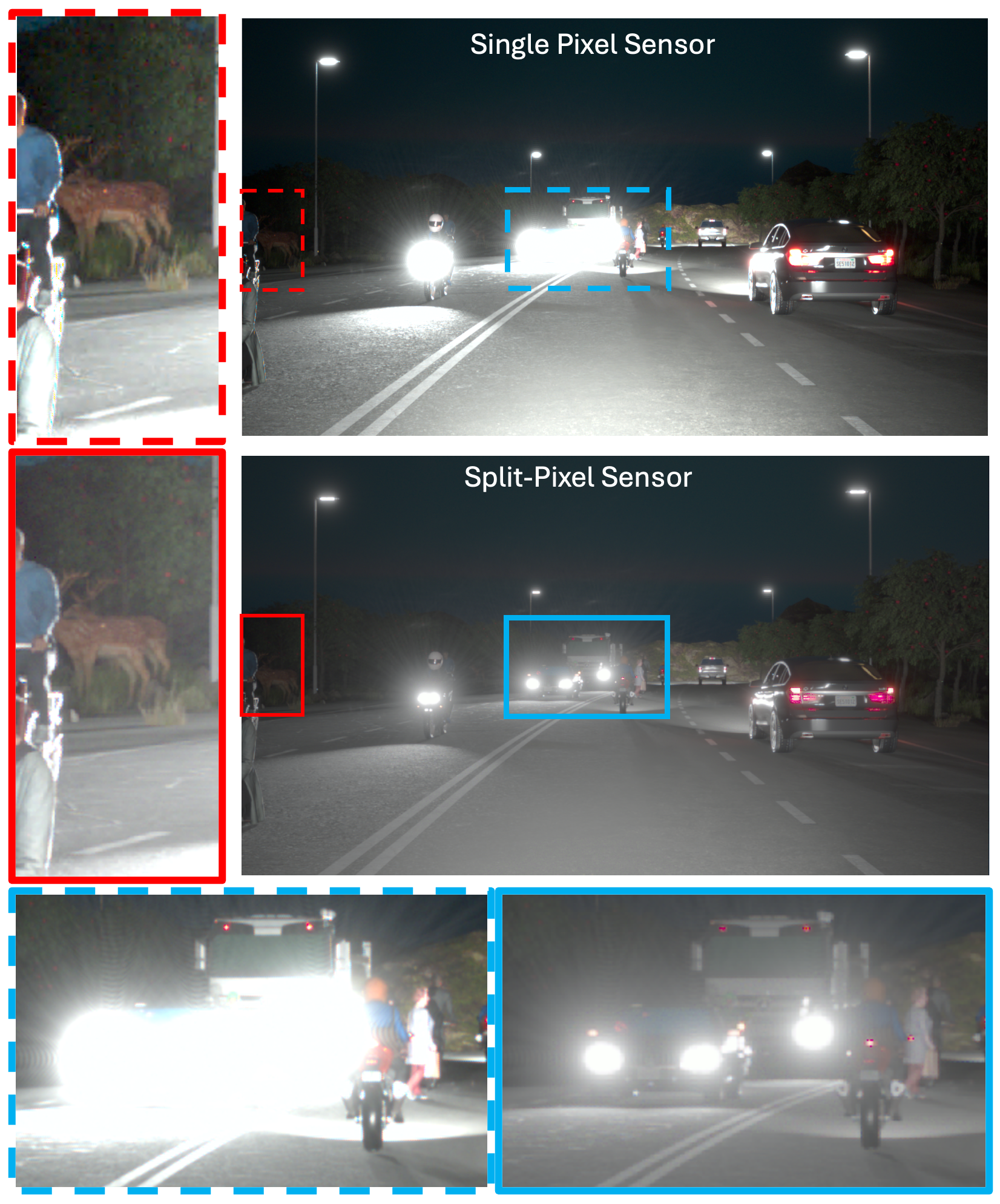}
    \caption{\textbf{A split pixel 3-capture sensor and flare.} We used the scene in Figure~\ref{fig:flareCompare} and to simulate the 3-capture split pixel image. The data from the large photodetector are saturated over large regions, due to lens flare (top). The small photodetector data preserve image contrast in parts of the image that are saturated by flare, and thus the combined image enhances the visibility of the motor cyclist (blue boxes). It combined image retains the visibility of the deer in the dark image region (red boxes).}
    \label{fig:splitFlare}
\end{figure}
Vendors have designed and built sensors to encode high dynamic range scenes in a single-shot. These sensors contain interleaved pixel arrays that are specialized to capture different luminance levels. One approach extends sensor dynamic range by reducing the sensitivity of a subset of interleaved pixels but retaining most of their well capacity. This enables the array to encode very bright regions of the scene that would normally saturate the pixel responses \cite{Nayar2000-spatially-varying}. This has been implemented in modern sensors by placing two photodetectors with different sizes in each pixel \cite{Willassen_2015_splitpixel, Innocent2019-nestedPD}. Such sensors are sometimes called split pixel because the photodetector within each pixel is split into two, unequal parts. It is also possible to read each photodetector twice, with different gains, which is called dual conversion gain. Such sensor systems are also called multi-capture because a single-shot acquires multiple captures from the interleaved pixel arrays \cite{Wandell1999-chiba-mcsi}. 
\begin{figure*}[t]
    \includegraphics[width=0.98\textwidth]{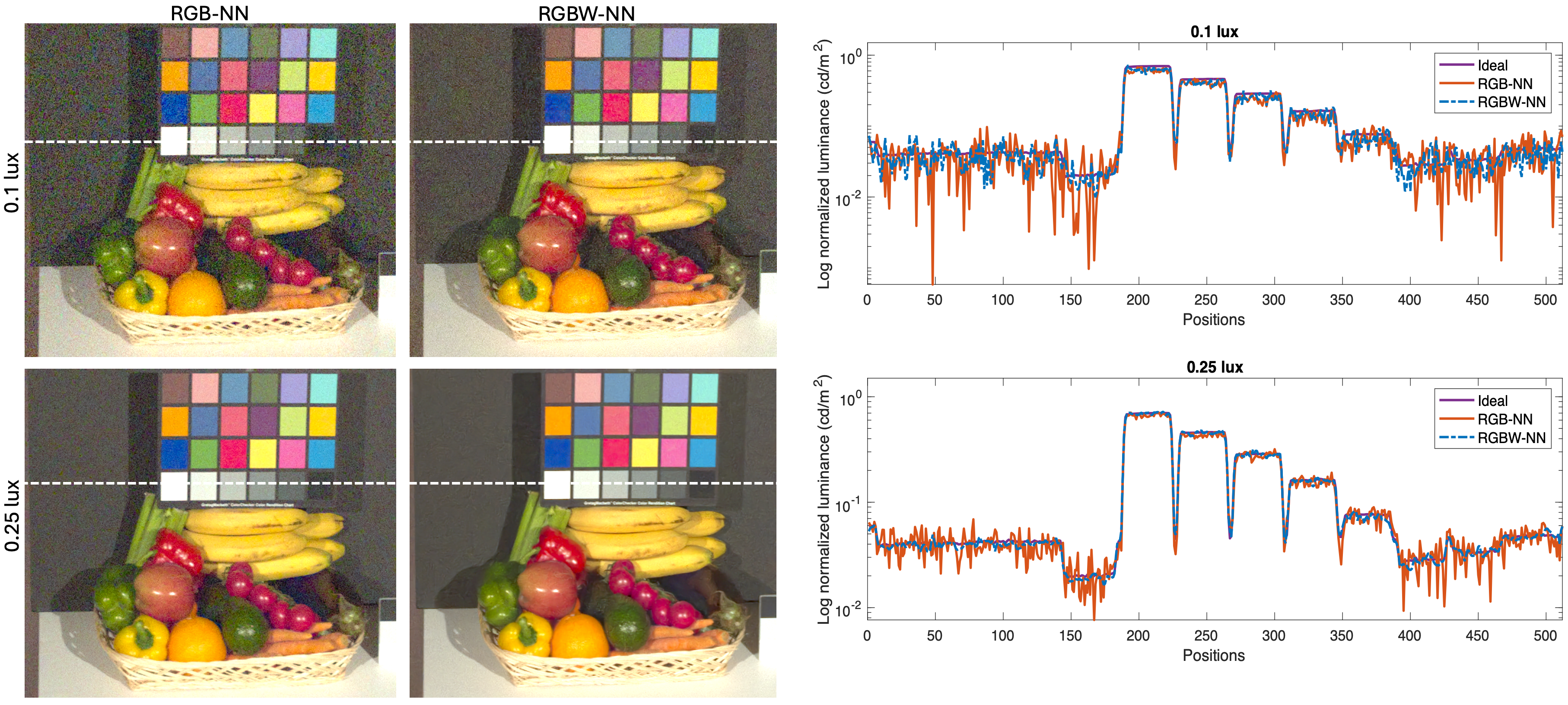}
    \caption{\textbf{Comparing sensors with RGB and RGBW color filter arrays}. \textit{Left}: The two rows simulate images illuminated at different mean levels (0.1 lux, 0.25 lux).  The two columns show simulations of RGB (left) and RGBW (right) sensor images reconstructed using trained restormer networks for demosaicing and denoising. The image quality of the RGBW reconstructions, particularly at darker illuminance levels, are better compared to RGB reconstructions. \textit{Right:} The graphs compare the reconstructed log luminance of the RGB, RGBW, and the ground truth (Ideal) measured along the dashed, white line. The RGBW reconstructions are superior in darker image regions. }
    \label{fig:CFADemosacing}
\end{figure*}

We simulated an Omnivision 3-capture split-pixel sensor comprising a large photodetector -- read with high and low gain -- and a small photodetector -- read once with low gain \cite{Solhusvik2019-split}. The small photodetector sensitivity is 100 times lower than that of the large photodetector. This design is advantageous for scenes that contain regions at very different mean luminance levels, such as the tunnel scene in Figure \ref{fig:splitTunnel}. 

The data from the three captures are integrated into a single final image as follows. The high and low gain images from the large pixel are combined by input-referring their values. When neither is saturated, the average is used; when the high gain is saturated, only the low gain is used. The large photodetector effectively captures the interior of the tunnel, but it saturates in the bright region near the tunnel exit. In this region, the data from the low-sensitivity small photodetector remain valid; these are input-referred and incorporated into the image. This approach to combining the three captures preserves image contrast both inside the tunnel and at the exit. Machine learning-based combination methods have also been explored in the literature \cite{robidoux2021end}.

Figure \ref{fig:splitFlare} analyzes a 3-capture sensor applied to the nighttime driving scene in Figure \ref{fig:flareCompare}. In the standard long duration capture, headlight flare obscures the motorcyclist. The split-pixel 3-capture sensor mitigates this issue by substituting saturated, large photodetector data within the flare region with corresponding data from the smaller photodetector, enhancing motorcyclist visibility. The deer, located in the darker area alongside the road, is captured by the large photodetector and is visible in both image versions.

The 3-capture sensor was simulated across the entire ISETHDR dataset. Object detection performance was compared between a standard sensor configuration (using only the LPLG detector) and the 3-capture approach. Measured this way, there were no significant performance gains for car detection (0.39 to 0.39) and person detection (0.224 to 0.244) using the split-pixel design. However, the clear benefits of this approach are evident in specific scenarios, such as the tunnel scene in Figure \ref{fig:splitTunnel} and flare scene in Figure~\ref{fig:splitFlare}.

\subsection{RGBW sensor}

\begin{figure}
    \centering
    \includegraphics[width=0.85\linewidth]{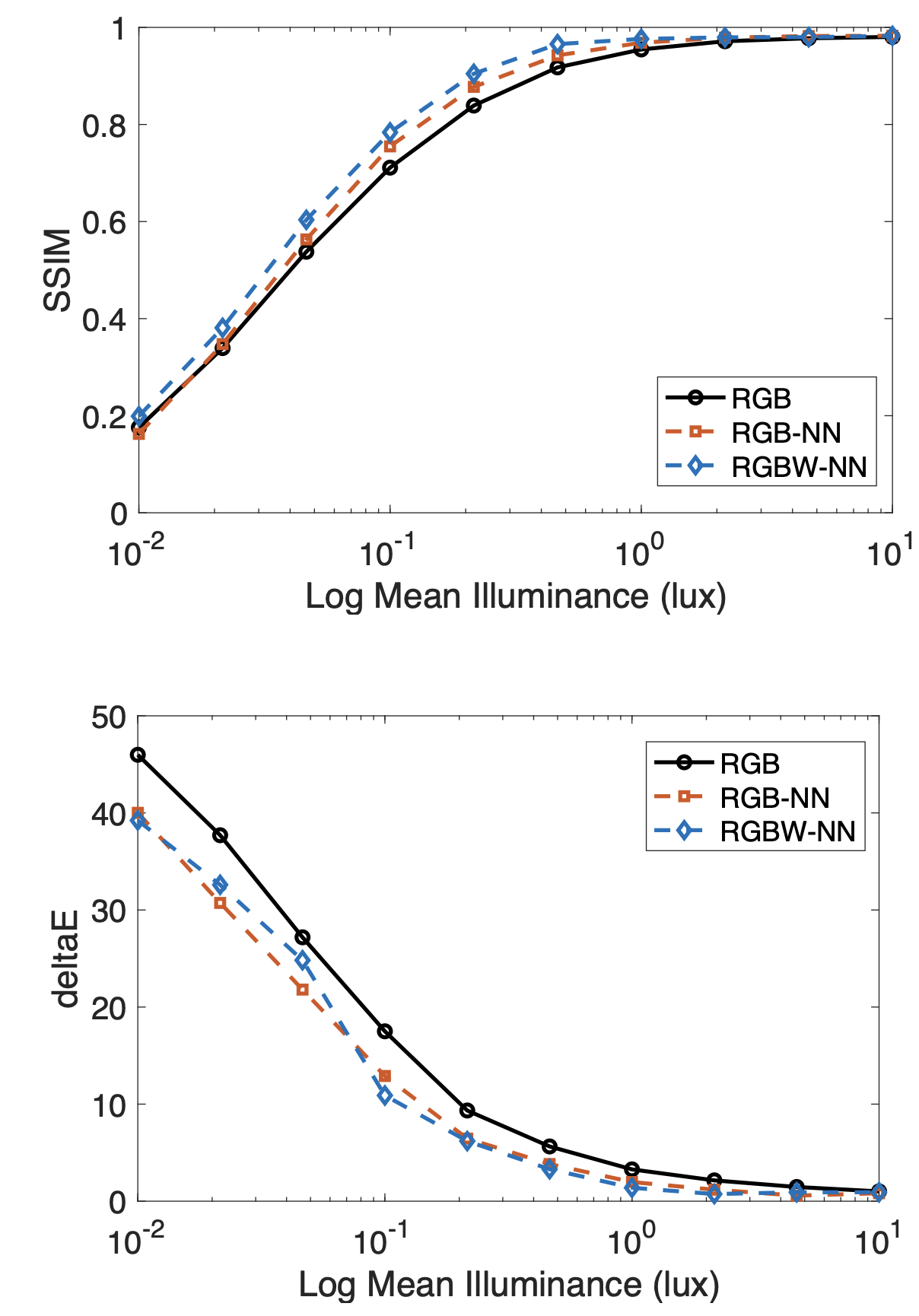}
    \caption{\textbf{RGBW Image quality}. The graphs show two measures of image quality (Structural Similarity Index Measure, SSIM; color accuracy of the Macbeth target, CIELAB $\Delta E$) across a wide range of mean illumination levels. The RGBW reconstruction achieves better image quality, particularly at lower mean levels, for both metrics. The results for a simulated RGB sensor with a conventional demosaicing (e.g. bilinear interpolation) and color reconstruction method is also shown for comparison. The neural networks both outperform the conventional method.}
    \label{fig:CFA_metrics}
\end{figure}
Sensor dynamic range can also be expanded by increasing the sensitivity of a subset of pixels in the sensor array. A typical Bayer RGB design has three color filters, each filter rejects roughly two-thirds of the incident photons. The G filter is present at twice the sampling density \cite{Bayer1976-CFApatent}; replacing one of the green filters, leaving the passage of light clear, creates an RGB-Clear, also called RGBW, array \cite{Parmar2009-RGBW}. The mosaic of clear pixels increases the system dynamic range because the clear pixels encode dark scene regions \cite{OPPO-RGBW,Vivo-RGBW,Canon-RGBW}.

However, there is a sensitivity mismatch between the three RGB pixels and the W pixel which has limited the commercial adoption of the RGBW sensor \cite{Kijima2007-RGBW}. In prior work, we used image systems simulation to develop and evaluate simple learning algorithms, based on regression, that leverage the advantage of the RGBW arrays \cite{jiang2017learning}. For this paper, we created ISETCam models of an RGB and an RGBW sensor with noise parameters that matched a commercial sensor from ON Semiconductor. The RGB color filters in the two sensor models are matched. We simulated sensor data to train restormer networks to demosaic and denoise the sensor inputs \cite{zamir2022restormer}. To train the networks, we simulated the sensor voltages of 13,103 realistic scenes combined with methods described, including scenes from the ISETHDR dataset as well as PBRT scenes designed with the book \cite{pbrt-bitterli}. The simulated scene illumination levels and dynamic ranges varied considerably; the training data included both dark and saturated regions. The scenes were processed through a diffraction-limited (f/\# 4) lens and then captured by sensors with either 1.5 or 3.0 microns. For each type of sensor, we also generated noise-free, fully sampled RGB and RGBW data that served as ground-truth. 

We trained one network to convert RGBW sensor data to ground-truth, and a second network to convert the Bayer RGB sensor data. To train the networks, we represented the noisy sensor data in four (RGBW) or three (RGB) channels, each at the spatial resolution of the final full sampled image. Missing values (e.g., the red, green and white values at a blue pixel) were filled with a small voltage sampled from a uniform distribution (0 - 0.001 volts) to prevent the network from over-fitting to the zero values. The networks map sensor mosaic data into a fully sampled,  noise-free, set of RGB  images, thus performing both demosaicing and denoising.

The RGBW sensor has better performance at low luminance levels, which can be seen from visual inspection in the dark regions of Figure \ref{fig:CFADemosacing}). The RGBW image is less noisy at low mean illuminance (0.1 lux), and the data are better in dark regions of the higher mean illuminance data (0.25 lux). We quantified the noise as a function of mean illumination level in Figure~\ref{fig:CFA_metrics}, using  SSIM and CIELAB $\Delta E$ metrics. These curves confirm that despite the sensitivity mismatch that has plagued RGBW systems, the neural network reconstructs images from the RGBW that are better at low mean illuminance levels and  equal to the RGB images at higher mean levels.


Finally, we evaluated object detection in the ISETHDR collection using the RGBW and RGB datasets using a COCO pretrained model (YOLOX) \cite{Lyu2022-objectdetectors}. The RGBW sensor outperformed the RGB sensor, achieving a mean Average Precision (mAP50) of 0.35 compared to 0.32 for the car class, using COCO metrics \cite{Lin2014-zb}. The superior RGBW image quality in the dark regions has a modest downstream benefit for object detection when measured with the ISETHDR nighttime driving dataset.

\section{Discussion} 
Advanced Driver Assistance Systems (ADAS) rely on camera data for multiple tasks, including road user identification, automatic emergency braking, lane departure warnings, and adaptive cruise control. Nighttime scenes are challenging to simulate and can cause problems for ADAS due to their high dynamic range and low illuminance level. Flare makes it difficult to accurately detect and interpret road users (cars, pedestrian, cyclists, etc.) and road conditions (Figures \ref{fig:flareCompare}, and \ref{fig:splitFlare}), resulting in reduced vehicle safety. Further, some ADAS systems, such as adaptive headlights or automatic high beams, are vulnerable to being triggered by the flare from other vehicles' headlights, leading to unintended activation or deactivation of these systems. 

A digital twin that quantitatively matches HDR nighttime driving scenarios, including scenarios with other road users, can help anticipate how the characteristics of the scene and imaging system may impact vehicle safety. The physics-based simulation we implemented includes quantitative optics and sensor models whose parameters match the parts deployed in automotive systems. Using explicit, realistic models of the image system components makes the simulation more useful for understanding and quantifying the impact of specific components \cite{Goossens-RTF, Farrell2012-digcamsimApOp}. The data and software presented in this paper are tools that synthesize such scenes for system evaluation and machine learning.

The digital twin simulations show that the split pixel sensor mitigates high dynamic range and flare problems (Figures \ref{fig:splitTunnel}) and \ref{fig:splitFlare}). These issues are partly what motivated the development of automotive LIDAR \cite{Lidar-BastosReview-2021} and gated-image sensors (\cite{Gated-Gruber-2021,Heide-Gated-2024}). Future evaluations of the value of these special systems should be compared with the performance obtained using split-pixel image sensors.

The end-to-end simulation enables us to identify and study specific conditions in which a sensor design proves valuable. The tunnel and flare scenes show the value of the split-pixel design, and low illuminance levels show the advantage of the RGBW design. Relying on the average performance of the ISETHDR collection, or any collection, can mask the value achieved in important cases. End-to-end simulation and digital twins enables a carefully tuned analysis of important conditions.

The simulations we describe can be improved and extended. The scattering flare model, while capturing key aspects of nighttime driving, requires further refinement to incorporate lens inter-reflection effects. To address the limitations of static scenes, we are developing tools to generate dynamic video sequences that simulate camera and object motion, enabling evaluation of time-varying sensor systems. Finally, we are working on tools to generate specific types of scenes on demand that can be used to evaluate important edge cases.

\section*{Acknowledgment}
We thank Devesh Upadhyay, and Alireza Rahimpour for their suggestions and encouragement in launching this project. We are grateful to Prof. Xiangyang Ji, Prof. Chuxi Yang, and Jiayue Xie for their support and assistance throughout the project. We thank Joyce E. Farrell for her thoughtful suggestions, encouragement, and careful reading of the manuscript. 

{\small
\UseRawInputEncoding
\bibliographystyle{ieeetr}
\bibliography{egbib}

\begin{thebibliography}{10}

\bibitem{Punnappurath2022-day2night}
A.~Punnappurath, A.~Abuolaim, A.~Abdelhamed, A.~Levinshtein, and M.~S. Brown,
  ``Day-to-night image synthesis for training nighttime neural {ISPs},'' in
  {\em 2022 {IEEE/CVF} Conference on Computer Vision and Pattern Recognition
  ({CVPR})}, IEEE, June 2022.

\bibitem{Dai2018-semanticsegDayNight}
D.~Dai and L.~Van~Gool, ``Dark model adaptation: Semantic image segmentation
  from daytime to nighttime,'' in {\em 2018 21st International Conference on
  Intelligent Transportation Systems ({ITSC})}, pp.~3819--3824, Nov. 2018.

\bibitem{Bui2020-pb}
H.~T.~T. Bui, D.~H. Le, T.~T.~A. Nguyen, and T.~V. Pham, ``Deep learning based
  semantic segmentation for nighttime image,'' 2020.

\bibitem{Sakaridis2022-DANighttimeSeg}
C.~Sakaridis, D.~Dai, and L.~Van~Gool, ``{Map-Guided} curriculum domain
  adaptation and {Uncertainty-Aware} evaluation for semantic nighttime image
  segmentation,'' 2022.

\bibitem{Xu2021-nightSegCurriculum}
Q.~Xu, Y.~Ma, J.~Wu, C.~Long, and X.~Huang, ``{CDAda}: A curriculum domain
  adaptation for nighttime semantic segmentation,'' 2021.

\bibitem{Wu2023-DANNet}
X.~Wu, Z.~Wu, L.~Ju, and S.~Wang, ``A {One-Stage} domain adaptation network
  with image alignment for unsupervised nighttime semantic segmentation,''
  2023.

\bibitem{Lin2021-d2n-IEEEIntellTransport}
C.-T. Lin, S.-W. Huang, Y.-Y. Wu, and S.-H. Lai, ``{GAN-Based} {Day-to-Night}
  image style transfer for nighttime vehicle detection,'' {\em IEEE Trans.
  Intell. Transp. Syst.}, vol.~22, pp.~951--963, Feb. 2021.

\bibitem{Zhu2017-cycleGAN}
J.-Y. Zhu, T.~Park, P.~Isola, and A.~A. Efros, ``Unpaired image-to-image
  translation using cycle-consistent adversarial networks,'' in {\em 2017
  {IEEE} International Conference on Computer Vision ({ICCV})}, pp.~2223--2232,
  IEEE, Oct. 2017.

\bibitem{Wang2023-nighttimedata}
X.~Wang, X.~Tu, B.~Al-Hassani, C.-W. Lin, and X.~Xu, ``Select informative
  samples for {Night-Time} vehicle detection benchmark in urban scenes,'' {\em
  Remote Sensing}, vol.~15, p.~4310, Aug. 2023.

\bibitem{Sakaridis2019-darkzurich}
C.~Sakaridis, D.~Dai, and L.~Gool, ``Guided curriculum model adaptation and
  uncertainty-aware evaluation for semantic nighttime image segmentation,''
  {\em ICCV}, pp.~7373--7382, Jan. 2019.

\bibitem{Sun2019-NightSemSeg-ZJUdata}
L.~Sun, K.~Wang, K.~Yang, and K.~Xiang, ``See clearer at night: towards robust
  nighttime semantic segmentation through day-night image conversion,'' in {\em
  Artificial Intelligence and Machine Learning in Defense Applications},
  vol.~11169, pp.~77--89, SPIE, Sept. 2019.

\bibitem{Yu2018-Berkeley-BDD}
F.~Yu, W.~Xian, Y.~Chen, F.~Liu, M.~Liao, V.~Madhavan, and T.~Darrell,
  ``{{BDD100K}}: A diverse driving video database with scalable annotation
  tooling,'' {\em arXiv [cs.CV]}, May 2018.

\bibitem{Wu2020-flare-bl}
Y.~Wu, Q.~He, T.~Xue, R.~Garg, J.~Chen, A.~Veeraraghavan, and J.~T. Barron,
  ``How to train neural networks for flare removal,'' {\em arXiv:2011.12485},
  Nov. 2020.

\bibitem{Dai2022-flare7k}
Y.~Dai, C.~Li, S.~Zhou, R.~Feng, and C.~C. Loy, ``{Flare7K}: A phenomenological
  nighttime flare removal dataset,'' {\em arXiv:2210.06570}, Oct. 2022.

\bibitem{Dai2023-flare7k++}
Y.~Dai, C.~Li, S.~Zhou, R.~Feng, Y.~Luo, and C.~C. Loy, ``{Flare7K++}: Mixing
  synthetic and real datasets for nighttime flare removal and beyond,'' {\em
  arXiv [cs.CV]}, June 2023.

\bibitem{Wu2019-night-rf}
C.-E. Wu, Y.-M. Chan, C.-H. Chen, W.-C. Chen, and C.-S. Chen, ``{IMMVP}: An
  efficient daytime and nighttime {On-Road} object detector,'' {\em
  arXiv:1910.06573}, Oct. 2019.

\bibitem{Qiao2021-flare}
X.~Qiao, G.~P. Hancke, and R.~W.~H. Lau, ``Light source guided single-image
  flare removal from unpaired data,'' in {\em 2021 {IEEE/CVF} International
  Conference on Computer Vision ({ICCV})}, pp.~4177--4185, IEEE, Oct. 2021.

\bibitem{Zhou2023-flareremoval-ICCV}
Y.~Zhou, D.~Liang, S.~Chen, S.~Huang, S.~Yang, and C.~Li, ``Improving lens
  flare removal with general-purpose pipeline and multiple light sources
  recovery,'' {\em ICCV}, pp.~12923--12933, Aug. 2023.

\bibitem{Farrell2012-digcamsimApOp}
J.~E. Farrell, P.~B. Catrysse, and B.~A. Wandell, ``Digital camera
  simulation,'' {\em Appl. Opt.}, vol.~51, pp.~A80--90, Feb. 2012.

\bibitem{Liu2019-softprototype-ICCV}
Z.~Liu, T.~Lian, J.~Farrell, and B.~Wandell, ``Soft prototyping camera designs
  for car detection based on a convolutional neural network,'' in {\em
  Proceedings of the {IEEE} International Conference on Computer Vision
  Workshops}, 2019.

\bibitem{pharr-book}
M.~Pharr, W.~Jakob, and G.~Humphreys, {\em Physically based rendering: From
  theory to implementation}.
\newblock Morgan Kaufmann, 2016.

\bibitem{lyu2021validation}
Z.~Lyu, K.~Kripakaran, M.~Furth, E.~Tang, B.~Wandell, and J.~Farrell,
  ``Validation of image systems simulation technology using a cornell box,''
  {\em arXiv preprint arXiv:2105.04106}, 2021.

\bibitem{Goossens-RTF}
T.~Goossens, Z.~Lyu, J.~Ko, G.~Wan, J.~Farrell, and B.~Wandell, ``Ray-transfer
  functions for camera simulation of 3d scenes with hidden lens design,'' 2022.

\bibitem{Farrell2008-sensorcalandsim}
J.~Farrell, M.~Okincha, and M.~Parmar, ``Sensor calibration and simulation,''
  in {\em Digital Photography {IV}}, vol.~6817, p.~68170R, International
  Society for Optics and Photonics, Mar. 2008.

\bibitem{Chen2009-isetvalidation}
J.~Chen, K.~Venkataraman, D.~Bakin, B.~Rodricks, R.~Gravelle, P.~Rao, and
  Y.~Ni, ``Digital camera imaging system simulation,'' {\em IEEE Trans.
  Electron Devices}, vol.~56, pp.~2496--2505, Nov. 2009.

\bibitem{iset3dsoftware}
Vistalab.stanford.edu, ``Iset3d.'' https://github.com/ISET/iset3d, 2022.

\bibitem{isetcamsoftware}
Vistalab.stanford.edu, ``Isetcam.'' https://github.com/ISET/isetcam, 2022.

\bibitem{roadrunner-mainpage}
``Roadrunner.'' \url{https://www.mathworks.com/products/roadrunner.html}, Mar.
  2023.
\newblock Accessed: 2023-2-28.

\bibitem{Liu2020-generalization}
Z.~Liu, T.~Lian, J.~Farrell, and B.~Wandell, ``Neural network generalization:
  The impact of camera parameters,'' {\em IEEE Access}, vol.~8,
  pp.~10443--10454, 2020.

\bibitem{Liu2021-depth-radiance}
Z.~Liu, J.~Farrell, and B.~A. Wandell, ``{ISETAuto}: Detecting vehicles with
  depth and radiance information,'' {\em IEEE Access}, vol.~9,
  pp.~41799--41808, 2021.

\bibitem{Lyu2022-validation}
Z.~Lyu, T.~Goossens, B.~A. Wandell, and J.~Farrell, ``Validation of
  {Physics-Based} image systems simulation with {3-D} scenes,'' {\em IEEE Sens.
  J.}, vol.~22, pp.~19400--19410, Oct. 2022.

\bibitem{Solhusvik2019-split}
J.~Solhusvik, T.~Willassen, S.~Mikkelsen, M.~Wilhelmsen, S.~Manabe, D.~Mao,
  Z.~He, K.~Mabuchi, and T.~Hasegawa, ``A {1280x960} 2.8$\mu$m {HDR} {CIS} with
  {DCG} and split-pixel combined,'' in {\em International Image Sensors
  Workshop}, 2019.

\bibitem{Willassen_2015_splitpixel}
T.~Willassen, J.~Solhusvik, R.~Johansson, S.~Yaghmai, H.~Rhodes, S.~Manabe,
  D.~Mao, Z.~Lin, D.~Yang, O.~Cellek, E.~Webster, S.~Ma, and B.~Zhang, ``A
  {1280x1080} $\mu$m split-diode pixel {HDR} sensor in 110nm {BSI} {CMOS}
  process,'' 2015.
\newblock Accessed: 2023-11-21.

\bibitem{Nayar2000-spatially-varying}
S.~K. Nayar and T.~Mitsunaga, ``High dynamic range imaging: spatially varying
  pixel exposures,'' in {\em Proceedings {IEEE} Conference on Computer Vision
  and Pattern Recognition. {CVPR} 2000 (Cat. {No.PR00662})}, vol.~1,
  pp.~472--479 vol.1, IEEE, 2000.

\bibitem{Innocent2019-nestedPD}
M.~Innocent, Ã.~D. Rodríguez, D.~Guruaribam, M.~Rahman, M.~Sulfridge,
  S.~Borthakur, B.~Gravelle, T.~Goto, N.~Dougherty, B.~Desjardin, D.~Sabo,
  M.~Mlinar, and T.~Geurts, ``Pixel with nested photo diodes and 120 {dB}
  single exposure dynamic range,'' in {\em International image sensors
  workshop}, pp.~95--98, 2019.

\bibitem{Wandell1999-chiba-mcsi}
B.~A. Wandell, P.~Catrysse, J.~M. DiCarlo, D.~Yang, and A.~El~Gamal, ``Multiple
  capture single image architecture with a {CMOS} sensor,'' in {\em Proc. Chiba
  Conf. Multispectral Imaging}, pp.~1--7, 1999.

\bibitem{robidoux2021end}
N.~Robidoux, L.~E.~G. Capel, D.-e. Seo, A.~Sharma, F.~Ariza, and F.~Heide,
  ``End-to-end high dynamic range camera pipeline optimization,'' in {\em
  Proceedings of the IEEE/CVF Conference on Computer Vision and Pattern
  Recognition}, pp.~6297--6307, 2021.

\bibitem{Bayer1976-CFApatent}
B.~Bayer, ``Color imaging array,'' 1976.

\bibitem{Parmar2009-RGBW}
M.~Parmar and B.~A. Wandell, ``Interleaved imaging: an imaging system design
  inspired by rod-cone vision,'' in {\em Digital Photography V} (B.~G. Rodricks
  and S.~E. Süsstrunk, eds.), SPIE, Jan. 2009.

\bibitem{OPPO-RGBW}
Oppo, ``{OPPO} unveils multiple innovative imaging technologies.''
  \url{https://tinyurl.com/yc8rz4nm}.
\newblock {Accessed: 2024-8-2}.

\bibitem{Vivo-RGBW}
Vivo, ``Vivo {X80} is the only vivo smartphone with a {Sony IMX866} sensor: The
  world's first {RGBW} bottom sensors.'' \url{https://tinyurl.com/ys22xkwr}.
\newblock {Accessed: 2024-8-2}.

\bibitem{Canon-RGBW}
Canon, ``{CANON} 19 pro {5G}.'' \url{https://tinyurl.com/mwssb3uv}.
\newblock {Accessed: 2024-8-2}.

\bibitem{Kijima2007-RGBW}
T.~Kijima, H.~Nakamura, J.~T. Compton, J.~F. Hamilton, and T.~E. DeWeese,
  ``Image sensor with improved light sensitivity,'' Nov. 2007.

\bibitem{jiang2017learning}
H.~Jiang, Q.~Tian, J.~Farrell, and B.~A. Wandell, ``Learning the image
  processing pipeline,'' {\em IEEE Transactions on Image Processing}, vol.~26,
  no.~10, pp.~5032--5042, 2017.

\bibitem{zamir2022restormer}
S.~W. Zamir, A.~Arora, S.~Khan, M.~Hayat, F.~S. Khan, and M.-H. Yang,
  ``Restormer: Efficient transformer for high-resolution image restoration,''
  in {\em Proceedings of the IEEE/CVF conference on computer vision and pattern
  recognition}, pp.~5728--5739, 2022.

\bibitem{pbrt-bitterli}
B.~Bitterli, ``Rendering resources,'' 2016.
\newblock https://benedikt-bitterli.me/resources/.

\bibitem{Lyu2022-objectdetectors}
C.~Lyu, W.~Zhang, H.~Huang, Y.~Zhou, Y.~Wang, Y.~Liu, S.~Zhang, and K.~Chen,
  ``{RTMDet}: An empirical study of designing real-time object detectors,''
  {\em arXiv [cs.CV]}, Dec. 2022.

\bibitem{Lin2014-zb}
T.-Y. Lin, M.~Maire, S.~Belongie, J.~Hays, P.~Perona, D.~Ramanan,
  P.~Doll{\'a}r, and C.~L. Zitnick, ``Microsoft {COCO}: Common objects in
  context,'' in {\em Computer Vision -- {ECCV} 2014}, Lecture notes in computer
  science, pp.~740--755, Cham: Springer International Publishing, 2014.

\bibitem{Lidar-BastosReview-2021}
D.~Bastos, P.~P. Monteiro, A.~S.~R. Oliveira, and M.~V. Drummond, ``An overview
  of {LiDAR} requirements and techniques for autonomous driving,'' in {\em 2021
  Telecoms Conference (ConfTELE)}, pp.~1--6, IEEE, Feb. 2021.

\bibitem{Gated-Gruber-2021}
T.~Gruber, S.~Walz, W.~Ritter, and K.~Dietmayer, ``High-resolution gated depth
  estimation for self-driving cars in {AdverseWeather}: Von der fahrerassistenz
  zum autonomen fahren 6. internationale {ATZ}-fachtagung,'' in {\em
  Automatisiertes Fahren 2020} (T.~Bertram, ed.), Proceedings, pp.~125--139,
  Wiesbaden: Springer Fachmedien Wiesbaden, 2021.

\bibitem{Heide-Gated-2024}
A.~Ramazzina, S.~Walz, P.~Dahal, M.~Bijelic, and F.~Heide, ``Gated fields:
  Learning scene reconstruction from gated videos,'' in {\em Proceedings of the
  IEEE/CVF Conference on Computer Vision and Pattern Recognition},
  pp.~10530--10541, 2024.

\end{thebibliography}
}

\end{document}